\documentclass[final,5p,times,twocolumn]{elsarticle}

\usepackage[T1]{fontenc}
\usepackage[utf8]{inputenc}
\usepackage{amsmath,amssymb,amsthm}
\usepackage{dsfont}
\usepackage{mathtools}
\usepackage{empheq}
\usepackage[binary-units]{siunitx}
\DeclareSIUnit\crate{C}
\usepackage{algorithm}
\usepackage{algpseudocode}
\usepackage{booktabs}
\usepackage{longtable}
\usepackage{doi}
\usepackage[capitalize, noabbrev]{cleveref}
\crefname{subsection}{Subsection}{Subsections}

\usepackage{hyperref}
\hypersetup{
  hidelinks=true
}
\usepackage{tikz}
\usetikzlibrary{calc}
\usepackage{pgfplots}
\usepackage{subcaption}
\usepackage{multirow}

\pgfplotsset{my style/.append style={axis x line=middle, axis y line=
		middle}}
\usepackage[printonlyused]{acronym}

\usepackage{multicol}
\usepackage{hyphenat}
\journal{Journal name}

\biboptions{sort&compress}

\newtheoremstyle{boldremark}
{\dimexpr\topsep/2\relax} 
{\dimexpr\topsep/2\relax} 
{}          
{}          
{\bfseries} 
{.}         
{.5em}      
{}          

\theoremstyle{boldremark}
\newtheorem{definition}{Definition}[section]

\newtheorem{problem}{Problem}[section]

\usepackage{xcolor}
\usepackage[normalem]{ulem}
\usepackage{cancel}
\newcommand{\stkout}[1]{\ifmmode\text{\sout{\ensuremath{#1}}}\else\sout{#1}\fi}

\begin{document}

  \begin{frontmatter}

    \title{Taper-based scattering formulation of the Helmholtz equation to improve the training process of Physics-Informed Neural Networks}

    \author[ianm]{W.~D\"orfler}

    \author[ianm]{M.~Elasmi\corref{correspondingauthor}}
	\ead{mehdi.elasmi@kit.edu}

    \author[ianm]{T.~Laufer\corref{correspondingauthor}}
    \cortext[correspondingauthor]{Corresponding authors}
    \ead{tim.laufer@kit.edu}

    \address[ianm]{Karlsruhe Institute of Technology~(KIT),
      Institute for Applied and Numerical Mathematics~(IANM),
      Englerstr.~2, 76131~Karlsruhe, Germany}

    \begin{abstract}
   	 This work addresses the scattering problem of an incident wave at a junction connecting two semi-infinite waveguides, which we intend to solve using \acp{PINN}. As with other deep learning-based approaches, \acp{PINN} are known to suffer from a spectral bias and from the hyperbolic nature of the Helmholtz equation. This makes the training process challenging, especially for higher wave numbers. We show an example where these limitations are present. In order to improve the learning capability of our model, we suggest an equivalent formulation of the Helmholtz \ac{BVP} that is based on splitting the total wave into a tapered continuation of the incoming wave and a remaining scattered wave. This allows the introduction of an inhomogeneity in the \ac{BVP}, leveraging the information transmitted during back-propagation, thus, enhancing and accelerating the training process of our \ac{PINN} model. The presented numerical illustrations are in accordance with the expected behavior, paving the way to a possible alternative approach to predicting scattering problems using \acp{PINN}. 
    \end{abstract}
    \acresetall
    

  \end{frontmatter}

  
  \section{Introduction}
  \label{sec:introduction}
  Since their introduction in $2019$ by Raissi et al.~\cite{RaissiKarniadakisInit}, \acp{PINN} have gained a huge popularity and interest in the scientific machine learning community and beyond. Remarkable results have been achieved across a wide range of engineering and physical problems. This involves studying both forward problems, e.g.,~\cite{raissi2020hidden,sahli2020physics,mao2020physics} and inverse problems, such as~\cite{jagtap2022physics,chen2020physics,depina2022application}. For a comprehensive review and discussion on \acp{PINN}, refer to~\cite{cuomo2022scientific}. 
  Despite being undeniably promising, \acp{PINN} still face several challenges, particularly for forward problems that exhibit higher frequencies, sharp transitions, and complex computational domains~\cite{Rahaman19,WANG2022110768,WANG2021113938}. Nevertheless, \acp{PINN} framework's flexibility has enabled the development of several extensions of the vanilla version that target various \ac{PINN}-related issues. One such extension is the \ac{SA-PINN}~\cite{McClenny2023}, which allows to adaptively correct possible discrepancies in the convergence rates for multiple-terms loss functions. Similarly,~\cite{WANG2022110768} studied and analyzed the latter problem using the \ac{NTK}, proposing a deterministic approach to compute suitable scalar weightings of the different loss function's components. In the same context of \ac{NTK}-theory,~\cite{WANG2021113938} introduced Fourier-feature networks to counteract the spectral bias of \acp{PINN}. On another note, domain decomposition approaches such as \acp{XPINN}~\cite{XPINN} provide a high potential framework to deal with large and complex computational domains.        
  In this work, we address the scattering problem of an incident wave at a junction connecting two semi-infinite closed waveguides. Such problems have been studied and analyzed using classical numerical techniques, both theoretically and numerically using the \ac{FEM}, for instance. An application can involve scattered elastic waves, however, we consider the simpler case of an optical wave. The underlying problem is governed by the Helmholtz equation, and a set of mixed boundary conditions using the \ac{DtN} operator. We refer, e.g., to~\cite{bourgeois2008linear,arens2008variational,ott2015domain}, for more details. 
  Solving the Helmholtz equation is challenging due to its hyperbolic nature and structure, even with classical iterative methods~\cite{ernst2011difficult,DIWAN2019110}, particularly for higher wave numbers. This challenge persists and is more pronounced when the solution framework is deep learning-based. To mitigate this, considering adaptive sine activation functions, hyper-parameter tuning, or extending the commonly considered fully-connected neural network by Fourier-feature mappings, enhances significantly the training capabilities of \acp{PINN}, see ~\cite{song2022versatile,ESCAPILINCHAUSPE2023126826,song2023simulating}, respectively. Furthermore, we refer to~\cite{schoder2024feasibility} for a study on the challenges and feasibility of solving the Helmholtz equation in $3$D.
  Perceiving the training difficulty of the considered Helmholtz \ac{BVP} with \acp{PINN}, in particular when no sources are present, we propose an equivalent taper-based scattering formulation that introduces inhomogeneities to the right-hand side of the Helmholtz equation and on the homogeneous Dirichlet boundary parts.
  We begin this article by introducing the classical formulation and the proposed taper-based scattering formulation of the Helmholtz \ac{BVP}. Next, we briefly define the considered \ac{PINN} model. In the penultimate section, we present the training results for both formulations using the same \ac{PINN} model for different wave numbers, and discuss our findings. Finally, we conclude the work with a brief summary and an outlook for possible improvements in future research. 
  \acresetall
  \section{The scattering problem}
  \label{sec:theory}
  In this section, we introduce the considered scattering problem of an incident wave at a two-dimensional waveguide junction.
  As illustrated in Figure~\ref{fig:WG}, the geometric setting consists of the junction $\Omega \subset \left(-b,b\right)\times \mathbb{R}, \, b>0$, connecting two straight semi-infinite closed waveguides, designated by $\Omega_-$ and $\Omega_+$. The waveguide junction is assumed to be bounded with boundary $\partial \Omega:= \Gamma_- \cup  \Gamma_{0,1} \cup \Gamma_{0,2} \cup \Gamma_+$. Thereby, $\Gamma_- := \bar{\Omega}_- \cap \bar{\Omega}$ and $\Gamma_+ := \bar{\Omega}_+ \cap \bar{\Omega}$ denote the interfaces between the waveguides and the junction, and constitute the fixed parts of the junction's shape, whereas $\Gamma_0:=\Gamma_{0,1} \cup \Gamma_{0,2}$ corresponds to the freely designable part. Note that the main geometric requirement that needs to be accounted for is guaranteeing a sufficiently regular boundary across the interfaces, e.g., with a $C^2$-smooth boundary. We refer to~\cite{ott2015domain} for more details and a thorough formulation of the considered framework. Given an appropriate incoming wave $u^\mathrm{inc}$ from $\Omega_-$ and propagating towards $\Omega$, we consider in this work the case of a closed waveguide junction that satisfies the Helmholtz \ac{BVP}, as defined in~\cite{ott2015domain}. First, the notion of a \ac{DtN} operator is required, see~\cite{bourgeois2008linear} for instance. \\ 
  \begin{minipage}{1\linewidth}
  	\medskip\centering 
  	\begin{tikzpicture}
	\begin{axis}[
		my style,
		xmin=-6, xmax=6, ymin=-5, ymax=5,
		xtick=\empty, 
		ytick=\empty, 
		xlabel=\empty, 
		ylabel=\empty, 
		axis x line=none, 
		axis y line=none, 
		]
		\node at (axis cs:0,1) {$\Omega$};
		\node at (axis cs:-4,1) {{$\Omega_{-}$}};
		\node at (axis cs:4,1) {{$\Omega_{+}$}};
		\addplot [dashed] coordinates { (-2,-1) (-2,3) } node[midway,left] {$\Gamma_-$};
		\addplot [dashed] coordinates { (2,0) (2,2) } node[midway,right] {$\Gamma_+$};
		\addplot[domain=-6:-2, color=black, line width=1.5pt]{3} node[midway,below] {};
		\addplot[domain=-6:-2, color=black, line width=1.5pt]{-1};
		\addplot[domain=2:6, color=black, line width=1.5pt]{2};
		\addplot[domain=2:6, color=black, line width=1.5pt]{0};
		\addplot[domain=-2:2, color=black, line width=1.5pt]{x*x*x/32-x*3/8+2.5} node[midway,above] {$\Gamma_{0,1}$};
		\addplot[domain=-2:2, color=black, line width=1.5pt]{-0.031*x*x*x*x-0.031*x*x*x+0.25*x*x+0.375*x-1} node[midway,below] {$\Gamma_{0,2}$};
		
		\node at (axis cs:-0.3,-4.2) (A) {};
		\node at (axis cs:0,-4.5) (D) {};
		\node at (axis cs:1.7,-4.2) (B) {$x$};
		\node at (axis cs:0,-2.5)  (C) {$z$};
		
		\draw[thick,->] (A) -- (B) {};
		\draw[thick,->] (D) -- (C) {};
		\node at (axis cs:-2,-2) []{$x=-b$};
		\node at (axis cs:2,-2) {$x=b$};
	\end{axis}

\end{tikzpicture}
  	
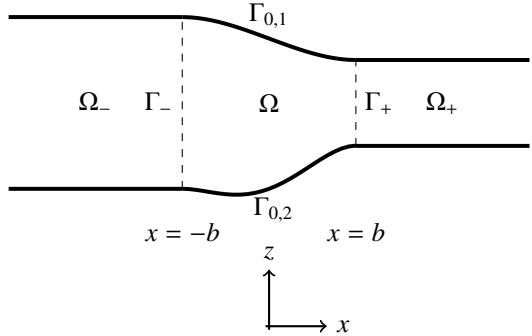
\captionof{figure}{Illustration of the considered type of waveguide junctions $\Omega$. Thereby, $\Omega_+$ and $\Omega_-$ denote two semi-infinite waveguides connected by $\Omega$ at the interfaces $\Gamma_-$ (at $x=-b$) and $\Gamma_+$ (at $x=b$), respectively. $\Gamma_{0,1}$ and $\Gamma_{0,2}$ designate the remaining boundaries of $\Omega$~\cite{ott2015domain}.}	\label{fig:WG} \medskip
  \end{minipage}%
  \begin{definition}[Dirichlet-to-Neumann (DtN) operator]\label{def:DtN}
  	Let $\mu_n^2 = n^2\pi^2$, $\phi_n\left(z\right) = \sqrt{2} \sin\left(n\pi z\right)$ for $z\in\left(0,1\right)$, and $k>0$. Provided $k^2 \neq \mu_n^2$ for all $n \in \mathbb{N}$, the \ac{DtN} operator is defined for $w \in L^2\left(\left(0,1\right)\right)$ by \[\Lambda w :=  w \mapsto \sum_{n \in \mathbb{N}} \iota \lambda_n \left(\phi_n,w\right)_{L^2(\left(0,1\right))}\phi_n.\]
  	Thereby, $\lambda_n := 
  	\begin{cases}
  		\sqrt{k^2-\mu^2_n}, & k^2 > \mu_n^2 \\
  		\iota\sqrt{\mu^2_n - k^2}, & k^2 \leq \mu_n^2
  	\end{cases}$ denotes the longitudinal frequency and $\iota$ the imaginary unit. 
  \end{definition}%
Then, the governing equations can be formulated as follows.	
  \begin{problem}[Classical Helmholtz \ac{BVP}] \label{prob:helmholtzBVP} Let $\Omega$ be a bounded waveguide junction as defined above with boundary $\Gamma:= \Gamma_{0,1} \cup \Gamma_{0,2} \cup \Gamma_+ \cup \Gamma_-$, and let $\nu$ denote a corresponding outer-pointing normal vector. For a wave number $k>0$, find $u \in \mathcal{C}^2\left(\Omega\right) \cap \mathcal{C}^1(\overline{\Omega})$, $(x,z) \mapsto u(x,z)$ such that
  	\begin{align*}
  			\Delta u + k^2 u &= 0 && \text{ in } \Omega,\\    
  			u &= 0   &&\text{ on } \Gamma_{0,1} \cup \Gamma_{0,2},\\
  			\partial_\nu u &= \Lambda u - 2\Lambda u^\text{inc} && \text{ on } \Gamma_-,\\
  			\partial_\nu u &= \Lambda u   &&  \text{ on } \Gamma_+.
  	\end{align*}  
  Thereby, $\partial_\nu u$ denotes the standard Neumann trace of $u$, and $\Lambda$ the \ac{DtN} operator. 
  \end{problem}%
	In addition, we propose the consideration of an equivalent formulation of \cref{prob:helmholtzBVP}, which we call the taper-based scattering formulation of the Helmholtz equation. The main idea of the ansatz consists in splitting the unknown wave function $u = u^\mathrm{sct}+\chi u^\mathrm{inc}$ to separate the incoming wave $u^\mathrm{inc}$ and rewrite the problem in terms of the scattered wave $u^\mathrm{sct}$. Thereby, we call the function $\chi: \left(-b,b\right) \to [0,1]$ a taper-function and define it as 
\begin{equation}
	\label{eq:chi}
	\chi(x) = 
	\begin{cases}
		\frac{-6}{b^5} x^5 - \frac{15}{b^4} x^4- \frac{10}{b^3} x^3, & x<0,\\
		0, & x\geq 0.
	\end{cases}
\end{equation}
Per construction, the taper-function satisfies $\chi(-b) = 1$ and $\chi(0) = 0$, and possesses vanishing first and second derivatives at $x = -b$ and $x = 0$ as well. Hence, $\chi \in \mathcal{C}^2((-b,b))$. With this and by taking into account that a suitable incoming wave $u^\text{inc}$ solves \cref{prob:helmholtzBVP}, the equivalent taper-based Helmholtz \ac{BVP} reads as follows.   
\begin{problem}[Taper-based scattering \ac{BVP}]\label{prob:taperBasedFormulation} For $k>0$ and $\chi$ as defined in \eqref{eq:chi}, find $u^\mathrm{sct} \in \mathcal{C}^2\left(\Omega\right) \cap \mathcal{C}^1(\overline{\Omega})$ such that
 \begin{align*}
		\Delta u^\mathrm{sct} + k^2 u^\mathrm{sct} &=
		-2\frac{\partial u^\text{inc}}{\partial x}\frac{\partial \chi}{\partial x}  -
		u^\text{inc}\frac{\partial^2 \chi}{\partial x^2} && \text{ in } \Omega,\\   
		u^\mathrm{sct} &= -u^\text{inc}\chi &&  \text{ on }\Gamma_{0,1} \cup \Gamma_{0,2},\\
		\partial_\nu u^\mathrm{sct}&=
		\Lambda u^\mathrm{sct} && \text{ on } \Gamma_- \cup \Gamma_+.
\end{align*}
\end{problem}
After solving \cref{prob:taperBasedFormulation}, the total wave $u$ is merely reconstructed by adding $u^\mathrm{inc}\chi$ to the solution $u^\mathrm{sct}$. Obviously, Problems~\ref{prob:helmholtzBVP} and \ref{prob:taperBasedFormulation} are equivalent, yet structurally different due to the inhomogeneity introduced to the BVP via the splitting. This is expected to improve the prediction capabilities of our model-based deep learning approach, namely \acp{PINN}, which we briefly introduce in the next section. 

  
  \section{Physics-Informed Neural Networks}
  \label{sec:PINN}
  
  \ac{PINN} was first introduced in 2019 by Raissi et al.~\cite{RaissiKarniadakisInit}. It consists in a hybrid approach that supplements a given \ac{ANN} by the underlying mathematical model (or parts of it), rendering a successful prediction physically consistent. In its vanilla version, the main building block of a \ac{PINN} is a deep \ac{ANN}, which consists of a plain stack of layers. The first and last layer are known as input and output layer, respectively, whereas the remaining ones correspond to the hidden layers. For $i=1,\dots,M$ and $M >1$, we denote the realization function of the $i$-th layer by $l_i: \mathbb{R}^{N_{i-1}} \to \mathbb{R}^{N_{i}}$, where $N_i$ is the number of neurons of the $i$-th layer. Formally, the network function $\mathrm{NN}_\theta$ of an \ac{ANN} mapping an input $x \in \mathbb{R}^{N_0}$ to an output $u_\theta := \mathrm{NN}_\theta(x) \in \mathbb{R}^{N_{M+1}}$ is defined recursively via 
  \begin{align*}
  	l_0 &:= x, \\
  	l_{i} &= \sigma_i \left(W_{i} l_{i-1} + b_{i}\right), \quad i = 1,\dots,M,\\
  	u_\theta &= W_{M+1} l_M + b_{M+1},
  \end{align*}
  where $\sigma_i$ is a possibly non-linear activation function for the $i$-th layer, and $W_i \in \mathbb{R}^{N_{i} \times N_{i-1}}$ and $b_{i} \in \mathbb{R}^{N_i}$ are known as the weight matrices and bias vectors, respectively. For convenience, let $\theta := (W_1,b_1,\dots,W_{M+1},b_{M+1})$ denote the set of all trainable parameters of the neural network.
	Using automatic differentiation~\cite{baydin2018a}, derivatives of the \ac{ANN}'s output function can be computed efficiently. This allows the penalization of non-physical solutions during the optimization process by incorporating the model equations into the loss function. For instance, let $\mathcal{L}$ be a differential operator (e.g., Helmholtz operator) and $\mathcal{B}_j$ some boundary operators, defining with suitable right-hand side $f$ and boundary conditions $g_j$ the following \ac{BVP}
  	\begin{equation}
  	\begin{aligned}	\label{problem:BVPgeneric}
  		\mathcal{L}[u]\left(x\right) &= f\left( x\right) && x \in \Omega, \\
  		\mathcal{B}_j[u]\left(x\right) &= g_j\left( x\right) && x \in \Gamma_j \subset \partial \Omega, \, j=1,\dots,4.
  	\end{aligned}
  	\end{equation}
  The \ac{PINN} ansatz consists in using $u_\theta := \mathrm{NN}_\theta(x)$ as a surrogate model for the solution $u$, cf.~\cite{RaissiKarniadakisInit}. In practice, training a \ac{PINN} model without additional data reduces to minimizing an empirical loss function that evaluates the model equations. Concretely, for the training points $x_\mathrm{r}= \{ x_{\mathrm{r},i}\}_{i=1}^{N_{\mathrm{r}}}  \subset \Omega$ and $x_{\mathrm{b}_j}= \{ x_{\mathrm{b}_j,i}\}_{i=1}^{N_{\mathrm{b}_j}}  \subset \Gamma_j,\,j=1,\dots,4,$ a typical loss function reads \[L(\theta,x_\mathrm{r},x_\mathrm{b}):= L_\mathrm{r}(\theta,x_\mathrm{r}) + \sum_{j=1}^{4}L_{\mathrm{b}_j}(\theta,x_{\mathrm{b}_j})\] with
  \begin{align*}
  	L_\mathrm{r}(\theta,x_\mathrm{r}) &= \frac{1}{N_\mathrm{r}}\sum_{i=1}^{N_\mathrm{r}}\left( \mathcal{L}[u_\theta]\left(x_{\mathrm{r},i}\right)-f\left(x_{\mathrm{r},i}\right) \right)^2 ,\\
  	L_{\mathrm{b}_j}(\theta,x_{\mathrm{b}_j}) &= \frac{1}{N_{\mathrm{b}_j}}\sum_{i=1}^{N_{\mathrm{b}_j}} \left( \mathcal{B}_j[u_\theta]\left(x_{\mathrm{b}_j,i}\right)-g_j\left(x_{\mathrm{b}_j,i}\right) \right)^2.
  \end{align*}
  The minimization procedure is usually performed using a gradient-descent type optimizer, e.g., \ac{Adam}~\cite{Kingma2014AdamAM}.
  As per all iterative methods, optimizing the \ac{ANN}'s trainable parameters require initialization. Popular choices include the Glorot~\cite{Glorot} or He initialization~\cite{He2015}, for instance. \\
  Although promising, the vanilla version of \ac{PINN} suffers from two main pathologies that limit its use cases. The first one, known as the spectral bias, is inherited from the conventional neural networks commonly used for \acp{PINN}. It states that neural networks tend to learn functions along the dominant eigendirections, which correspond to the lowest frequencies. In other words, they struggle to learn functions with higher frequencies. We refer the reader, e.g., to~\cite{Rahaman19,WANG2021113938} and the references therein for more details. The second pathology addresses a possible imbalance in the convergence rates of the constituent terms of the total loss function, which could result in neglecting certain aspects of the solution, hence to training failure, see ~\cite{WANG2022110768,Heydari2019SoftAdaptTF}, for instance. Several extensions of \ac{PINN} that address and (partially) circumvent these effects have been proposed and successfully deployed. However, to the authors knowledge, there is still no generic \ac{PINN} approach that allows efficient predictions of forward problems based solely on the mathematical model. In this work, we consider the \ac{SA-PINN}, introduced in~\cite{McClenny2023}. This approach is based on soft attention mechanisms and allows the reduction of convergence discrepancies of the loss terms by introducing a self-adaptive mask function $m: \left[0,\infty\right)\rightarrow \left[0,\infty\right)$ that associates to each training point a trainable weight. The function $m$ is assumed to be differentiable and strictly increasing, see~\cite{McClenny2023} for possible choices of $m$ and their effect on the training procedure. Provided $\lambda_\mathrm{r} = \left(\lambda_{\mathrm{r},1}, \dots, \lambda_{\mathrm{r},N_\mathrm{r}} \right)$ and $\lambda_\mathrm{b}:=\left(\lambda_{\mathrm{b}_j} \right)_{j=1}^4$ with $\lambda_{\mathrm{b}_j} = \left(\lambda_{\mathrm{b}_j,1}, \dots, \lambda_{\mathrm{b},N_{\mathrm{b}_j}} \right)$ are non-negative, the extended loss function, reads
  \begin{equation}
  		\label{eq:totalLoss}
  		L\left( \theta,x_\mathrm{r},x_\mathrm{b},\lambda_{\mathrm{r}},\lambda_{\mathrm{b}}\right) = L_{\mathrm{r}}\left( \theta,x_\mathrm{r},\lambda_{\mathrm{r}}\right)
  		+ \sum_{j=1}^{4} L_{\mathrm{b}_j}\left( \theta,x_{\mathrm{b}_j},\lambda_{\mathrm{b}_j}\right),
  \end{equation}
 where 
  \begin{align*}
  	L_\mathrm{r}\left( \theta,x_\mathrm{r},\lambda_\mathrm{r}\right) &= \frac{1}{N_\mathrm{r}}\sum_{i=1}^{N_\mathrm{r}}m\left(\lambda_{\mathrm{r},i}\right) \left( \mathcal{L}[\theta]\left(x_{\mathrm{r},i}\right)-f\left(x_{\mathrm{r},i}\right) \right)^2,\\
  	L_{\mathrm{b}_j}\left( \theta,x_{\mathrm{b}_j},\lambda_{\mathrm{b}_j}\right) &= \frac{1}{N_{\mathrm{b}_j}}\sum_{i=1}^{N_{\mathrm{b}_j}}m\left(\lambda_{\mathrm{b}_j,i}\right) \left( \mathcal{B}_j[\theta]\left(x_{\mathrm{b}_j,i}\right)-g_j\left(x_{\mathrm{b}_j,i}\right) \right)^2.
  \end{align*}
  With this, the optimization problem turns to a saddle point problem that simultaneously minimizes $\theta$, and penalizes the regions with higher losses by maximizing $\lambda_\mathrm{r}$ and $\lambda_\mathrm{b}$, i.e., solving
  \begin{equation}
  	\label{eq:saddlePointProblem}
  	\min_{\theta} \max_{\lambda_\mathrm{r},\lambda_\mathrm{b}} L\left( \theta,\lambda_\mathrm{r},\lambda_\mathrm{b}\right).
  \end{equation}
Besides \ac{SA-PINN}, we consider layer-wise adaptive activation functions to further ameliorate the approximation capabilities of our \ac{PINN} model~\cite{JAGTAP2020109136}. In particular, we use for all $i = 1,\dots,M$, $\sigma_i: x\mapsto \tanh(\alpha_i x)$, where $(\alpha_i)_i$ are trainable coefficients to be minimized along with the network weights and biases. In a slight abuse of notation, we include $(\alpha_i)_i$ in the set of trainable parameters $\theta$. 
 \section{Numerical Studies}
 \label{sec:results}
  \begin{figure*}[!ht]
 	\centering
 	\begin{subfigure}[b]{0.45\textwidth}
 		\centering 
 		\subfloat[Successfull prediction for $k=8$.]{
\begin{tikzpicture}[scale=0.85]

\definecolor{darkgray176}{RGB}{176,176,176}

\begin{axis}[
colorbar,
colorbar style={ytick={-1,0.0,1},yticklabels={
	\(\displaystyle -1\),	
  \(\displaystyle 0\),
\(\displaystyle 1\),
},ylabel={}},
colormap/viridis,
point meta max=1.00,
point meta min=-1.,
tick align=outside,
tick pos=left,
title={},
x grid style={darkgray176},
xmin=-2, xmax=2,
xtick style={color=black},
xtick={-2,-1.5,-1,-0.5,0,0.5,1,1.5,2},
axis equal image,
axis equal image,
xticklabels={
  \(\displaystyle -2.0\),
  \(\displaystyle -1.5\),
  \(\displaystyle -1.0\),
  \(\displaystyle -0.5\),
  \(\displaystyle 0.0\),
  \(\displaystyle 0.5\),
  \(\displaystyle 1.0\),
  \(\displaystyle 1.5\),
  \(\displaystyle 2.0\)
},
y grid style={darkgray176},
ymin=0, ymax=1,
ytick style={color=black},
ytick={0,0.5,1},
yticklabels={\(\displaystyle 0.0\),\(\displaystyle 0.5\),\(\displaystyle 1.0\)}
]
\addplot graphics [includegraphics cmd=\pgfimage,xmin=-2, xmax=2, ymin=0, ymax=1] {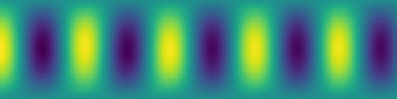};
\end{axis}

\end{tikzpicture}}{\label{k8}}
 	\end{subfigure}
 	\hfill
 	\begin{subfigure}[b]{0.45\textwidth}
 		\centering
 		\subfloat[Failed prediction of the expected amplitude for $k=13$.]{
\begin{tikzpicture}[scale=0.85]

\definecolor{darkgray176}{RGB}{176,176,176}

\begin{axis}[
colorbar,
colorbar style={ytick={-0.4,0.1,0.6},yticklabels={
  \(\displaystyle -0.4\),
  \(\displaystyle 0.1\),
  \(\displaystyle 0.6\)
},ylabel={}},
colormap/viridis,
point meta max=0.614904463291168,
point meta min=-0.454231679439545,
tick align=outside,
tick pos=left,
title={},
axis equal image,
x grid style={darkgray176},
xmin=-2, xmax=2,
xtick style={color=black},
xtick={-2,-1.5,-1,-0.5,0,0.5,1,1.5,2},
xticklabels={
  \(\displaystyle -2.0\),
  \(\displaystyle -1.5\),
  \(\displaystyle -1.0\),
  \(\displaystyle -0.5\),
  \(\displaystyle 0.0\),
  \(\displaystyle 0.5\),
  \(\displaystyle 1.0\),
  \(\displaystyle 1.5\),
  \(\displaystyle 2.0\)
},
y grid style={darkgray176},
ymin=0, ymax=1,
ytick style={color=black},
ytick={0,0.5,1},
yticklabels={\(\displaystyle 0.0\),\(\displaystyle 0.5\),\(\displaystyle 1.0\)}
]
\addplot graphics [includegraphics cmd=\pgfimage,xmin=-2, xmax=2, ymin=0, ymax=1] {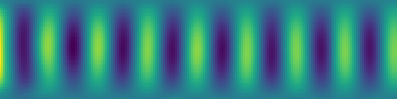};
\end{axis}

\end{tikzpicture}}{\label{k13}}
 	\end{subfigure}
 	\caption{Visualization of the real part of the PINN's prediction $\hat{u}:= u_\theta$ using \cref{prob:helmholtzBVP} after $50000$ \ac{Adam} training steps.}
 	\label{fig:HelmholtzPlots}
 \end{figure*} %
 \begin{figure*}[th!]
 	\centering
 	\begin{subfigure}[b]{0.45\textwidth}
 		\centering
 		\subfloat[Successfull prediction for $k=13$.]{
\begin{tikzpicture}[scale=0.85]

\definecolor{darkgray176}{RGB}{176,176,176}

\begin{axis}[
colorbar,
colorbar style={ytick={-1,0.0,1},yticklabels={
		\(\displaystyle -1\),	
		\(\displaystyle 0\),
		\(\displaystyle 1\),
	},ylabel={}},
colormap/viridis,
point meta max=1.00,
point meta min=-1.,
tick align=outside,
tick pos=left,
title={},
x grid style={darkgray176},
xmin=-2, xmax=2,
xtick style={color=black},
xtick={-2,-1.5,-1,-0.5,0,0.5,1,1.5,2},
axis equal image,
axis equal image,
xticklabels={
	\(\displaystyle -2.0\),
	\(\displaystyle -1.5\),
	\(\displaystyle -1.0\),
	\(\displaystyle -0.5\),
	\(\displaystyle 0.0\),
	\(\displaystyle 0.5\),
	\(\displaystyle 1.0\),
	\(\displaystyle 1.5\),
	\(\displaystyle 2.0\)
},
y grid style={darkgray176},
ymin=0, ymax=1,
ytick style={color=black},
ytick={0,0.5,1},
yticklabels={\(\displaystyle 0.0\),\(\displaystyle 0.5\),\(\displaystyle 1.0\)}
]
\addplot graphics [includegraphics cmd=\pgfimage,xmin=-2.01005029678345, xmax=2.01005029678345, ymin=-0.0128205129876733, ymax=1.01282048225403] {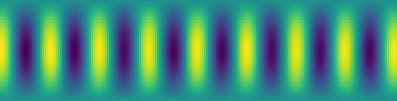};
\end{axis}

\end{tikzpicture}}{\label{k13taper}}
 	\end{subfigure}
 	\hfill
 	\begin{subfigure}[b]{0.45\textwidth}
 		\centering
 		\subfloat[Successfull prediction for $k=16$.]{
\begin{tikzpicture}[scale=0.85]
	
	\definecolor{darkgray176}{RGB}{176,176,176}
	
	\begin{axis}[
		colorbar,
		colorbar style={ytick={-1,0.0,1},yticklabels={
				\(\displaystyle -1\),	
				\(\displaystyle 0\),
				\(\displaystyle 1\),
			},ylabel={}},
		colormap/viridis,
		point meta max=1.00,
		point meta min=-1.,
		tick align=outside,
		tick pos=left,
		title={},
		x grid style={darkgray176},
		xmin=-2, xmax=2,
		xtick style={color=black},
		xtick={-2,-1.5,-1,-0.5,0,0.5,1,1.5,2},
		axis equal image,
		axis equal image,
		xticklabels={
			\(\displaystyle -2.0\),
			\(\displaystyle -1.5\),
			\(\displaystyle -1.0\),
			\(\displaystyle -0.5\),
			\(\displaystyle 0.0\),
			\(\displaystyle 0.5\),
			\(\displaystyle 1.0\),
			\(\displaystyle 1.5\),
			\(\displaystyle 2.0\)
		},
		y grid style={darkgray176},
		ymin=0, ymax=1,
		ytick style={color=black},
		ytick={0,0.5,1},
		yticklabels={\(\displaystyle 0.0\),\(\displaystyle 0.5\),\(\displaystyle 1.0\)}
		]
		\addplot graphics [includegraphics cmd=\pgfimage,xmin=-2.01005029678345, xmax=2.01005029678345, ymin=-0.0128205129876733, ymax=1.01282048225403] {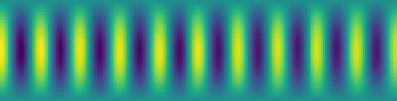};
	\end{axis}
	
\end{tikzpicture}}{\label{k16taper}}
 	\end{subfigure}
 	\begin{subfigure}[b]{0.45\textwidth}
 		\centering
 		\subfloat[Tapered incoming wave $\chi u^\mathrm{inc}$ for $k=13$.]{
\begin{tikzpicture}[scale=0.85]

\definecolor{darkgray176}{RGB}{176,176,176}

\begin{axis}[
colorbar,
colorbar style={ytick={-1,0.0,1},yticklabels={
		\(\displaystyle -1\),	
		\(\displaystyle 0\),
		\(\displaystyle 1\),
	},ylabel={}},
colormap/viridis,
point meta max=1.00,
point meta min=-1.,
tick align=outside,
tick pos=left,
title={},
x grid style={darkgray176},
xmin=-2, xmax=2,
xtick style={color=black},
xtick={-2,-1.5,-1,-0.5,0,0.5,1,1.5,2},
axis equal image,
axis equal image,
xticklabels={
	\(\displaystyle -2.0\),
	\(\displaystyle -1.5\),
	\(\displaystyle -1.0\),
	\(\displaystyle -0.5\),
	\(\displaystyle 0.0\),
	\(\displaystyle 0.5\),
	\(\displaystyle 1.0\),
	\(\displaystyle 1.5\),
	\(\displaystyle 2.0\)
},
y grid style={darkgray176},
ymin=0, ymax=1,
ytick style={color=black},
ytick={0,0.5,1},
yticklabels={\(\displaystyle 0.0\),\(\displaystyle 0.5\),\(\displaystyle 1.0\)}
]
\addplot graphics [includegraphics cmd=\pgfimage,xmin=-2.01005029678345, xmax=2.01005029678345, ymin=-0.0128205129876733, ymax=1.01282048225403] {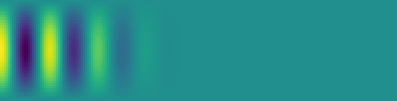};
\end{axis}

\end{tikzpicture}}{\label{k13taperedInput}}
 	\end{subfigure}
 	\hfill
 	\begin{subfigure}[b]{0.45\textwidth}
 		\centering
 		\subfloat[\ac{PINN} prediction of the scattering wave $u_\theta:= u^\mathrm{sct}$ for $k=13$.]{
\begin{tikzpicture}[scale=0.85]

\definecolor{darkgray176}{RGB}{176,176,176}

\begin{axis}[
colorbar,
colorbar style={ytick={-1,0.0,1},yticklabels={
		\(\displaystyle -1\),	
		\(\displaystyle 0\),
		\(\displaystyle 1\),
	},ylabel={}},
colormap/viridis,
point meta max=1.00,
point meta min=-1.,
tick align=outside,
tick pos=left,
title={},
x grid style={darkgray176},
xmin=-2, xmax=2,
xtick style={color=black},
xtick={-2,-1.5,-1,-0.5,0,0.5,1,1.5,2},
axis equal image,
axis equal image,
xticklabels={
	\(\displaystyle -2.0\),
	\(\displaystyle -1.5\),
	\(\displaystyle -1.0\),
	\(\displaystyle -0.5\),
	\(\displaystyle 0.0\),
	\(\displaystyle 0.5\),
	\(\displaystyle 1.0\),
	\(\displaystyle 1.5\),
	\(\displaystyle 2.0\)
},
y grid style={darkgray176},
ymin=0, ymax=1,
ytick style={color=black},
ytick={0,0.5,1},
yticklabels={\(\displaystyle 0.0\),\(\displaystyle 0.5\),\(\displaystyle 1.0\)}
]
\addplot graphics [includegraphics cmd=\pgfimage,xmin=-2.01005029678345, xmax=2.01005029678345, ymin=-0.0128205129876733, ymax=1.01282048225403] {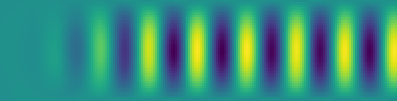};
\end{axis}

\end{tikzpicture}}{\label{k13NNoutput}}
 	\end{subfigure}
 	\caption{Visualization of the real part of the reconstructed solution $\hat{u}:=u_\theta + \chi u^\mathrm{inc}$ and its constituents using \cref{prob:taperBasedFormulation} after $50000$ \ac{Adam} training steps.}
 	\label{fig:HelmholtzPlotsTaper}
 \end{figure*} %
 In this section, we compare the approximation capability of the previously discussed \ac{PINN} model when applied to Problems~\ref{prob:helmholtzBVP}~and~\ref{prob:taperBasedFormulation}. For the sake of comparability, we use the same network architecture and hyper parametrization for both cases. In particular, the employed \ac{ANN} consists of $M = 10$ hidden layers with $N_i = 45$ neurons, for all $i=1,\dots,M$. As mentioned in the previous section, we consider adaptive hyperbolic tangents as activation functions initialized with $\alpha_i=2$, for $i= 1,\dots,M$, whereas the remaining trainable parameters of $\theta$ are initialized using a Glorot normal initializer. \\
 
 For the sake of simplicity, we consider a rectangular waveguide junction $\Omega := \left(-b,b\right)\times\left(0,1\right)$, $b=2$. With the boundaries of $\Omega$, $\Gamma_{0,1}$, $\Gamma_{0,2}$, $\Gamma_-$, and $\Gamma_+$ corresponding to $\Gamma_j$, $j=1,\dots,4$ in \eqref{problem:BVPgeneric}, respectively, and by assigning accordingly the right-hand sides, we arrive at a loss function for Problems~\ref{prob:helmholtzBVP} and \ref{prob:taperBasedFormulation} similar to \eqref{eq:totalLoss}. Thereby, the interior and boundary points $x_\mathrm{r}$ and $x_{\mathrm{b}_j}$ are taken as a uniform grid of size $(120,10)$, and with $N_{\mathrm{b}_j}:=N_\mathrm{b}=80$ for all $j=1,\dots,4$, respectively. Moreover, we define the considered self-adaptive mask function by $m:x \mapsto x^2$, and draw the initial self-adaptive weights from uniform distributions as $\lambda_\mathrm{r} \sim \left(\mathcal{U}(0,0.5)\right)^{N_\mathrm{r}}$, $\lambda_{b_j} \sim \left(\mathcal{U}(0,30) \right)^{N_\mathrm{b}},$ for $j=1,2$, and $\lambda_{b_j}\sim \left(\mathcal{U}(0,10)\right)^{N_\mathrm{b}}$, for $j=3,4$. 
 To solve \eqref{eq:saddlePointProblem}, we employ an \ac{Adam} optimizer with an exponential learning rate decay (with a rate of $0.95$ every $1000$ steps). With this, a training process consists of $50000$ steps. All computations are performed using a TensorFlow v2.11.0 implementation in Python v3.10.12. To ensure reproducibility of the results, we choose the number $11$ as seed for the generation of pseudo-random numbers. \\
 
 In the following experiments, we assume that only the first mode propagates through the waveguide, i.e., in \cref{def:DtN}, we assume $n=1$. Moreover, the incoming wave reads \[u^{\mathrm{inc}}\left(x,z\right) =  \frac{1}{\sqrt{2}}e^{\iota(x+b)\cdot \lambda_1}\phi_1(z), \quad (x,z) \in \Omega.\]
 Note that in the considered case of a simple rectangular waveguide junction, the energy transmission is total, hence the expected exact solution $u_\mathrm{ref}$ corresponds to the incoming wave propagating along the $x$-axis. \\
 On one hand, we notice that the prediction results of our \ac{PINN} model using the ansatz $\hat{u} :=u_\theta = u$ in \cref{prob:helmholtzBVP}, i.e, when applied to the classical Helmholtz \ac{BVP} are only successful up to $k=8$. In~\cref{fig:HelmholtzPlots}\subref{k8}, we illustrate the real part of the predicted solution for a successful example at $k=8$. By increasing the wave number, we observe that the \ac{PINN} solution starts to degenerate. In particular, the predicted amplitude decreases with higher $k$. For example, we showcase this effect for $k=13$ in \cref{fig:HelmholtzPlots}\subref{k13}. On the other hand, the application of the same model on the equivalent formulation with $u_\theta = u^\mathrm{sct}$ in \cref{prob:taperBasedFormulation} seems to clearly ameliorate the approximation capacity in dependence of $k$. This is visualized in \cref{fig:HelmholtzPlotsTaper}. In particular, \cref{fig:HelmholtzPlotsTaper}\subref{k13taper} shows the reconstructed solution $\hat{u}:=u_\theta + \chi u^\mathrm{inc}$ for $k=13$, which is in agreement with the expected solution. Moreover, we verify that this improvement is maintained up to $k=16$, see, e.g., \cref{fig:HelmholtzPlotsTaper}\subref{k16taper}. For convenience, the constituents of $\hat{u}$, namely, the \ac{PINN} prediction $u_\theta$ and $\chi u^\mathrm{inc}$ are plotted in \cref{fig:HelmholtzPlotsTaper}\subref{k13taperedInput} and \cref{fig:HelmholtzPlotsTaper}\subref{k13NNoutput}, respectively. Furthermore, we compare in Table~\ref{tab:comparisonErrors} the prediction results for different wave numbers $k$ using both formulations. For this, we use as metric the relative error with $\|\cdot\|_2$ being the Euclidean norm
 \begin{equation}\label{eq:relError}
 	\varepsilon(u):= \frac{\| u - u_\mathrm{ref} \|_2}{\|u_\mathrm{ref}\|_2},
 \end{equation}
and denote the \ac{PINN}-based solution again by $\hat{u}$. Note that the latter is understood as $\hat{u}:= u_\theta$ for \cref{prob:helmholtzBVP} and as $\hat{u} = u_\theta + \chi u^\mathrm{inc}$ for \cref{prob:taperBasedFormulation}.  \\
\begin{minipage}{1\linewidth}\medskip
	\begin{center}
		\begin{tabular}{c|c|c|c|c|c|c|c} 
			& $k$& 8 & 9 & 10 & 13 & 15 & 16 \\
			\hline
			\multirow{ 2}{*}{Prob.~\ref{prob:helmholtzBVP}} & $\varepsilon_\mathrm{R}$ & $0.057$ &$0.158$ & $0.21$& $0.58$& $-$& $-$ \\ 
			&$\varepsilon_\mathrm{I}$ & $0.059$ &$0.15$ &$0.22$&$0.6$& $-$& $-$\\
			\hline \hline
			\multirow{ 2}{*}{Prob.~\ref{prob:taperBasedFormulation}} & $\varepsilon_\mathrm{R}$ & $0.018$ & $0.024$& $0.028$& $0.036$&$0.071$& $0.09$\\ 
			&$\varepsilon_\mathrm{I}$ & $0.012$ &$0.023$& $0.038$& $0.045$&$0.089$& $0.1$ \\
			\hline
		\end{tabular}
		\captionof{table}{Relative errors $\varepsilon_\mathrm{R}:=\varepsilon(\Re\{\hat{u}\})$ and $\varepsilon_\mathrm{I}:=\varepsilon(\Im\{\hat{u}\})$ w.r.t.~the real and imaginary parts of $\hat{u}$, respectively (denoted by $\Re\{\hat{u}\}$ and $\Im\{\hat{u}\}$) for Problems~\ref{prob:helmholtzBVP} and \ref{prob:taperBasedFormulation} at different $k$.}\label{tab:comparisonErrors}
	\end{center}
	\smallskip
\end{minipage} %
The accuracy of the trained models is computed for $k = 8,9,10,13,15,16$. We notice the expected amelioration even for $k=8$, where the classical formulation in \cref{prob:helmholtzBVP} is sufficient to successfully learn the solution in the waveguide junction. However, it should also be noted that the formulation of \cref{prob:taperBasedFormulation} does not completely circumvent the spectral bias but reduces it. This can be seen from the increasing error in both cases with increasing wave numbers $k$. Besides extending the range of possible wave numbers, the considered equivalent formulation accelerates the training procedures, as the same accuracy is reached at a significantly earlier stage of the training. For instance, \cref{fig:err_vs_iter} shows the evolution of the training error for the solution's real part $\Re\{\hat{u}\}$ w.r.t.~the training iterations at $k=8$, where both an improved and more accurate learning can be distinguished.
\begin{minipage}{1\linewidth}\bigskip
	\centering
	\input{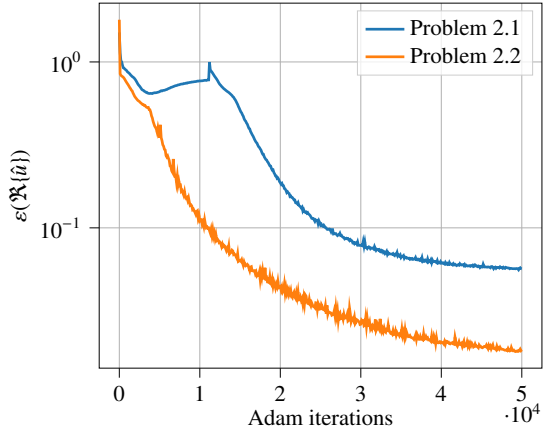}
	\captionof{figure}{Comparison of the learning dynamics using the relative error \eqref{eq:relError} w.r.t.~the real part of the predicted solution $\hat{u}$ (denoted by $\Re\{\hat{u}\}$) from Problems~\ref{prob:helmholtzBVP} and \ref{prob:taperBasedFormulation} with $k=8$. The relative error is evaluated for visualization purpose once each $100$ iterations.} \label{fig:err_vs_iter}\medskip
\end{minipage} %
  
  \section{Summary and Conclusion}
  \label{sec:conclusion}
  We proposed an equivalent taper-based scattering formulation of the two-dimensional Helmholtz \ac{BVP} by introducing inhomogeneities to the \ac{BVP} through a splitting of the solution. We numerically illustrated that the resulting equivalent formulation improves and accelerates the learning process of our \ac{PINN} model, as it reduces the tendency of our model when applied to the classical formulation of the Helmholtz \ac{BVP}, to learn the zero function. This allowed for more accurate solutions and extended the range of successfully predictable wave numbers.
  Although our numerical experiments were limited to straight waveguide junctions, it is possible to consider more complex geometrical configurations. Moreover, it should be noted that other shapes for the taper-function are conceivable, including non-polynomial functions. In addition, the taper-based scattering formulation is not restricted to the first mode, allowing for future extensions to consider higher modes for the incoming wave and the \ac{DtN} operator.
  
  \section*{Declaration of competing interest}

  \noindent
  The authors declare that they have no known competing financial interests or
  personal relationships that could have appeared to influence the work in this
  paper.

  %

  \section*{Acknowledgement}

  \noindent
  W.~Dörfler gratefully acknowledges the support of the Deutsche Forschungsgemeinschaft (DFG) within the SFB 1173 ``Wave Phenomena'' (Project-ID 258734477).\\
  M.~Elasmi and T.~Ļaufer acknowledge financial support by the German Research
  Foundation~(DFG) through the Research Training Group 2218
  SiMET~--~Simulation of Mechano-Electro-Thermal processes in Lithium-ion
  Batteries, project number 281041241. \\
  The authors want to thank M.~Sukhova (KIT) for her contributions in this work at early stages. Additionally, the authors want to thank R.~Schoof (KIT) for his general support.


	
	
	\appendix

	\addcontentsline{toc}{section}{Appendices}

	\section*{List of abbreviations}
	\label{app:abbreviations_and_symbols}
	\vspace*{-0.5cm}
	\subsection*{}
	\begin{acronym}[SA-PINN]
		\setlength{\itemsep}{0pt}
		\acro{Adam}{Adaptive moment estimation}
		\acro{ANN}{Artificial Neural Network}
		\acro{BVP}{Boundary Value Problem}
		\acro{DtN}{Dirichlet-to-Neumann}
		\acro{FBPINN}{Finite Basis Physics-Informed Neural Network}
		\acro{FEM}{Finite Element Method}
		\acro{PDE}{Partial Differential Equation}
		\acro{PINN}{Physics-Informed Neural Network}
		\acro{SA-PINN}{Self-Adaptive Physics-Informed Neural Network}
		\acro{tanh}{tangent hyperbolicus}
		\acro{XPINN}{Extended Physics-Informed Neural Network}
		\acro{NTK}{Neural Tangent Kernel}
		\acro{$hp$-VPINNs}{$hp$-Variational Physics Informed Neural Network}
	\end{acronym}

  \bibliography{literature}
\end{document}